\PassOptionsToPackage{table}{xcolor}
\documentclass[runningheads]{llncs}

\usepackage{eccv}

\usepackage{eccvabbrv}

\usepackage{graphicx}
\usepackage{booktabs}

\usepackage[accsupp]{axessibility}  

\usepackage[pagebackref,breaklinks,colorlinks]{hyperref}

\usepackage{orcidlink}

\usepackage{multirow}
\usepackage{makecell}
\usepackage{tikz}
\usetikzlibrary{shapes.geometric}
\usetikzlibrary{fit}
\usetikzlibrary{decorations.pathreplacing}
\usetikzlibrary{positioning, shapes, calc, arrows, arrows.meta, backgrounds, shadows, matrix}
\usepackage{arydshln} 
\usepackage{booktabs}
\usepackage{pifont}
\usepackage{algpseudocode,algorithm}

\usepackage{pgfplots}
\pgfplotsset{compat=1.17}
\usepackage{hhline}
\definecolor{lightgray}{gray}{0.95}
\begin{document}

\title{GPT4SGG: Synthesizing Scene Graphs from Holistic and Region-specific Narratives} 

\titlerunning{GPT4SGG}

\author{Zuyao Chen\inst{1,2} \and
Jinlin Wu\inst{2,3}  \and
Zhen Lei\inst{2,3} \and 
Zhaoxiang Zhang \inst{2, 3} \and 
Changwen Chen \inst{1}
}
\authorrunning{Z. Chen et al.}
%
\institute{The Hong Kong Polytechnic University  \\ 
\and 
Centre for Artificial Intelligence and Robotics, HKISI, CAS \and 
NLPR, Institute of Automation, Chinese Academy of Sciences, Beijing, China
\\
\email{zuyao.chen@connect.polyu.hk} \\ 
\email{\{jinlin.wu, zlei\}@nlpr.ia.ac.cn, zhaoxiang.zhang@ia.ac.cn} 
\email{changwen.chen@polyu.edu.hk}
}
\maketitle

\begin{abstract}
Training Scene Graph Generation (SGG) models with natural language captions has become increasingly popular due to the abundant, cost-effective, and open-world generalization supervision signals that natural language offers.
However, such unstructured caption data and its processing pose significant challenges in learning accurate and comprehensive scene graphs.
The challenges can be summarized as three aspects: 
1) traditional scene graph parsers based on linguistic representation often fail to extract meaningful relationship triplets from caption data. 
2) grounding unlocalized objects of parsed triplets will meet ambiguity issues in visual-language alignment. 
3) caption data typically are sparse and exhibit bias to partial observations of image content.
Aiming to address these problems,  we propose a divide-and-conquer strategy with a novel framework
named \textit{GPT4SGG}, to obtain more accurate and comprehensive scene graph signals. 
This framework decomposes a complex scene into a bunch of simple regions, 
resulting in a set of region-specific narratives. 
With these region-specific narratives (partial observations) and a holistic narrative (global observation) for an image, 
a large language model (LLM) performs the relationship reasoning to synthesize 
an accurate and comprehensive scene graph.
Experimental results demonstrate \textit{GPT4SGG} significantly improves the performance of SGG models trained on image-caption data,
in which the ambiguity issue and long-tail bias have been well-handled with more accurate and comprehensive scene graphs. 

\end{abstract}

\input{pics/fig_intro}
\section{Introduction}
\label{sec:intro}
Scene Graph Generation (SGG) offers a visual symbolic representation,
where nodes encode object categories and spatial information,
and edges define the spatial relationships and interactions between objects.
Conventional SGG models \cite{xu2017scene,zellers2018neural,tang2019learning,tang2020unbiased,chiou2021recovering,li2021bipartite} rely on datasets annotated manually, encompassing objects and their inter-relations.
For an image with $N$ objects, there are $N*(N-1)$ potential pairs for annotation, making the creation of large-scale datasets both time-consuming and costly. 
Recent studies \cite{zhong2021learning,li2022integrating,zhang2023learning,chen2023expanding} leverage image-caption data to learn scene graphs from natural language description directly,
which is commonly referred to as \textit{language-supervised SGG} or \textit{weakly-supervised SGG}. 
Under this setting, 
a traditional NLP tool, scene parser for SGG \cite{mao2018parser}, is employed to extract relationship triplets (\ie, $<subject, predicate, object>$) from unstructured textual data. 
The parsed triplets with grounded objects serve as the pseudo label for training SGG models.

Nevertheless, several inherent challenges hinder the effective learning of scene graphs from natural language descriptions, as shown in Fig. \ref{fig:challenges}.
\textbf{1)} The scene parser \cite{mao2018parser} struggles to extract complete scene graphs from caption data, failing to derive relationship triplets even from simple sentences like \textit{a small car is parked in front of a scooter}. 
For instance, it has been observed that there are 8\% failure cases of the scene parser \cite{mao2018parser} on COCO caption \cite{chen2015microsoft} train set, highlighting its difficulty in accurately interpreting and structuring relationship information from natural language descriptions.
By contrast, GPT-4 \cite{gpt4} works well for this task. However, directly replacing the scene parser with GPT-4 does not address the following issues. 
\textbf{2)} Ambiguity arises in grounding unlocalized objects of parsed triplets through visual-language alignment, particularly when multiple instances of the same category appear in an image, making it challenging to accurately match visual regions with language queries. 
\textbf{3)} Caption data, even with manual annotations, exhibit sparsity and bias, often emphasizing specific observations while overlooking crucial visual cues for generating comprehensive scene graphs.
These challenges are intertwined and significantly hamper learning scene graphs from natural language, requiring more effective solutions.

\input{pics/fig_pipeline}
\label{sec:pipe}
In contrast to conventional methods \cite{zhong2021learning,li2022integrating,zhang2023learning,chen2023expanding} , our framework, \textit{GPT4SGG}, innovates by reversing the pipeline of language-supervised SGG, as shown in \cref{fig:pipeline}.
Previous approaches typically follow the pipeline:
(1) \textit{parsing relationship triplets from caption data} $\rightarrow$ (2) \textit{grounding unlocalized objects of parsed triplets} $\rightarrow$ (3) \textit{learning scene graphs with parsed triplets and grounded objects.}
However, this pipeline faces significant challenges discussed before.
For instance, given a parsed triplet <\textit{person}, \textit{on},
\textit{tie}>, it is very difficult to choose which object is our target only based on the word ``\textit{person}'' and the visual input, although the grounding model knows there are a man and a woman in this image. 
Our framework addresses these issues by:
\begin{enumerate}
    \item (\textcolor{ourPipelineColor}{Step 1: Grounding}) Prioritizing the localization of objects using annotations or object detectors, ensuring accurate visual-language alignment ;
    \item  (\textcolor{ourPipelineColor}{Step 2: Decomposing}) Decomposing a complex scene into a set of regions.
    Specifically,  both holistic and region-specific narratives 
    are generated to inform the relationship reasoning process, thereby mitigating caption data biases ;
    \item  (\textcolor{ourPipelineColor}{Step 3: Relation Reasoning}) Utilizing localized objects and a set of observations (narratives) as input for LLMs like GPT-4, allowing for more precise relationship deduction between objects.
\end{enumerate}
By utilizing this divide-and-conquer strategy,  \textit{GPT4SGG} substantially enhances the accuracy and comprehensiveness of the generated scene graphs, as our experiments demonstrate.

To validate the proposed framework, 
we build two SGG-aware instruction-following datasets and
conduct extensive experiments with state-of-the-art SGG models. 
The first instruction-following dataset, ``COCO@GPT'', 
is built on the COCO detection dataset \cite{coco2014}. 
The second dataset, ``VG@GPT'',
is built on the VG150 dataset \cite{xu2017scene}.
We only included images where the number of objects is two or more. 
This selection is crucial to ensure the dataset is sufficiently complex and representative for boosting LLMs to synthesize scene graphs in instruction-following scenarios.

Considering the privacy and limitations of OpenAI, we fine-tune a private LLM, Llama 2 \cite{touvron2023llama}. 
Llama 2 \cite{touvron2023llama}, an open-source LLM comparable to GPT-3, along with Alpaca \cite{taori2023stanford}, Vicuna \cite{chiang2023vicuna}, use machine-generated instructions to enhance alignment capabilities, showing remarkable performance.
The instruction tuning process follows Alpaca \cite{taori2023stanford} with the instruction-following data generated by GPT-4 \cite{gpt4}.

In short, our contributions are as follows
\begin{itemize}
    \renewcommand{\labelitemi}{\textbullet}
    \item We introduce \textit{GPT4SGG}, a novel framework that leverages the capabilities of LLMs, particularly GPT-4, for Scene Graph Generation. 
    To our knowledge, this is a pioneer work in this field.
    \item  We develop two specialized instruction-following datasets derived from the COCO detection dataset and VG150, aiming to evaluate and enhance LLMs' SGG abilities in complex visual contexts.
    \item We fine-tune a private and SGG-aware LLM, Llama 2, using instruction-following data generated by GPT-4.
    \item Extensive experiments with state-of-the-art SGG models validate the effectiveness of \textit{GPT4SGG}, highlighting its capability to generate more accurate and comprehensive scene graphs.
\end{itemize}
\section{Related Work}
\textbf{Scene Graph Generation (SGG)} aims to create a structured representation of objects and their relationships within an image. 
Johnson \textit{et al.} \cite{jonhson2015irsg} first introduced the concept of scene graphs, presenting them as a method to improve image retrieval and scene understanding. 
The evolution of SGG has been marked by an increasing focus on extracting more detailed and contextually rich information from visual content.
This field has expanded significantly with contributions from both vision-based and language-based methodologies, providing a comprehensive understanding of scene composition and dynamics.
Earlier approaches to SGG predominantly leveraged deep learning models to identify and classify objects and their interrelations within images. 
These methods, such as IMP \cite{xu2017scene}, 
MOTIFS \cite{zellers2018neural}, primarily depended on convolutional neural networks (CNNs) or recurrent neural networks (RNNs) to extract visual features. 
However, the reliance on manually annotated data limits the scope of relationship recognition.

Recent works \cite{zhong2021learning,zhang2023learning,chen2023expanding} integrate natural language processing into SGG, providing a cheap yet valuable scheme for capturing complex relationships. 
These models could learn rich relationship concepts without manual annotations by employing image-caption pairs. 
This approach, while effective in certain contexts, faced limitations in scalability and the diversity of relationship types it could capture.
Previous methods of learning scene graphs from image-caption data are mainly struggled by the three challenges discussed in Sec. \ref{sec:intro}. 
This work reverses previous pipeline and presents a novel framework to address these challenges.

\noindent\textbf{Large Language Model (LLM)} has gained increasing attention on many complex reasoning tasks, including multi-hop question answering \cite{LanHJ0ZW21}, multi-turn conversation \cite{baize2023}, program synthesis \cite{nijkamp2022codegen}, \textit{etc.}  
Among a series of LLMs, GPT-4 \cite{gpt4}, developed by OpenAI, presents a remarkable capability to solve complex reasoning tasks. As reported in \cite{sparks2023}, GPT-4 can solve many novel and challenging tasks such as mathematics, medicine, finance, law, etc.  
These applications showcase the great potentiality of GPT-4 with task-specific prompts. 
Beyond textual data, the recently released GPT-4V \cite{OpenAI2023GPT4V} can access image input and perform various visual reasoning tasks. 
Nevertheless, the vision model of GPT-4V remains like a black box, with its API currently limited in speed and usage (as of February 2024).

Beyond GPT-4, the open-sourced Llama2 \cite{touvron2023llama} 
 and its variants have emerged for various applications, 
 especially for vision tasks. 
For instance,  Mini GPT-4 \cite{minigpt2023} utilizes Vicuna \cite{chiang2023vicuna} and a frozen vision encoder to learn a multi-modality alignment, 
enabling the generation of detailed image descriptions, website creation from sketches, and other complex reasoning tasks;
LLaVA \cite{llava2023} leverages visual instruction tuning with language-only GPT-4 and a frozen vision encoder, enabling advanced multi-modal understanding and interaction capabilities. 
Regardless of these attempts, the SGG capability of LLMs still needs to be explored. 
In this work, we will explore how to generate an accurate and comprehensive scene graph with the help of LLMs. 
\\

\noindent\textbf{Concurrent Work.} Parallel to our efforts, the LLM4SGG framework \cite{kim2023llm4sgg} utilizes an LLM to extract triplets from original and paraphrased captions. 
Despite its advantage over traditional scene parsers like \cite{mao2018parser}, 
it still follows the pipeline discussed in \cref{sec:pipe}.
Thus, it cannot avoid the ambiguity issue in grounding unlocalized objects. 
Moreover, relying on paraphrasing captions still cannot obtain 
comprehensive scene graphs for complex scenes. 
\section{Method}
This section will describe the framework, \textit{GPT4SGG}, how to generate more accurate and comprehensive scene graphs than the previous pipeline.  
An overview of \textit{GPT4SGG} is illustrated in Fig. \ref{fig:gpt4sgg}. 
\subsection{Object Grounding}
Object detection or grounding have been well-studied, such as 
\cite{ren2015faster,redmon2016you,detr2020,zareian2021open,li2022grounded,li2022grounded,liu2023grounding}. 
Briefly, we can use manual annotation or object detectors to obtain 
large-scale images with localized objects. 
This prior step ensures that each visual object is accurately aligned with a category or a phrase, preventing the hard choice when multiple instances of the same category exist.    
\input{pics/fig_gpt_sgg}

\subsection{Scene Decomposing: Holistic \& Region Narratives Generation}
\label{sec:narratives}
\makeatletter
\algrenewcommand\alglinenumber[1]{\footnotesize #1}
\makeatother

\begin{algorithm}[t]
\scriptsize
\centering
\begin{minipage}{.95\textwidth}
\caption{Generating Holistic and Region-specific Narratives}
\begin{algorithmic}[1]
\Require Original Image, N (maximum number of RoIs)
\Statex

\Procedure{GenerateNarratives}{Image}
    \State GlobalDescription $\gets$ \Call{Blip-2}{Image}
    \State Objects $\gets$ \Call{Detector}{Image} or \Call{Annotation}{Image}
    \State RoIList $\gets$ \Call{SelectRoIs}{Objects, N}
    \State LocalisedDescriptions $\gets$ [ ]
    \For{each RoI in RoIList}
        \State CroppedRegion $\gets$ Crop Region with RoI 
        \State Description $\gets$ \Call{Blip-2}{CroppedRegion}
        \State \textbf{Add} Description to LocalisedDescriptions
    \EndFor
    \State \textbf{return} GlobalDescription, LocalisedDescriptions
\EndProcedure
\item[]
\Function{SelectRoIs}{Objects, N}
    \State ValidPairs $\gets$ [ ]
    \State ObjectPairs $\gets$ pairwise-combinations(Objects)
    \For{pair in ObjectPairs}
        \If{IoU(pair) $>$ 0}
            \State \textbf{Add} pair to ValidPairs
        \EndIf
    \EndFor
    \State Shuffle(ValidPairs)
    \State ValidPairs $\gets$ ValidPairs[:N]
    \State RoIList $\gets$ Union of all pairs in ValidPairs
    \State \textbf{return} RoIList
\EndFunction

\end{algorithmic}
\label{algo: narratives}
\end{minipage}
\end{algorithm}

``\textit{A picture is worth a thousand words}''.
An Image of a complex scene has many plentiful clues and is hard to be summarized with short sentences by captioning models.
An intuitive idea is to ``look'' an image multiple times, like using sliding windows in R-CNN \cite{girshick2015region}.
However, with the localized objects,  we can reduce the 
complexity by selecting representative Region-of-Interest (RoI) areas.

The construction process is shown in Algorithm \ref{algo: narratives}.
Specifically, the holistic narrative is generated by feeding the whole image to the captioning model, providing a holistic description of the image content ; 
the region-specific narratives are produced by captioning a set of RoI areas, focusing on the content of local regions, especially the interaction between two objects. 
To select meaningful and representative RoIs for an input image, 
Localized objects are pairwise combined and filtered with an Intersection-over-Union (IoU) greater than zero. 
This selection criterion for RoI pairs is grounded in the premise that objects with a non-zero IoU are more likely to have a meaningful relationship as indicated in \cite{zellers2018neural,zhong2021learning}, which also reduces the computation burden for captioning and complexity for LLM's inference. 
By leveraging object categories, spatial locations, holistic narratives, and region-specific descriptions, LLMs can ``see'' images and interpret images without direct visual data input, facilitating complex visual reasoning tasks like SGG.

Another advantage of this textual representation for images is 
that captioning models pre-trained on  large-scale datasets typically receive a relatively small input resolution like $224\times 224$ in BLIP-2 \cite{blip2}, while 
this textual representation is independent of image resolution.
For images with high-resolution and complex scenes,
we can obtain plentiful and meaningful descriptions focusing on interactions between two objects.

\subsection{Relation Reasoning: Synthesizing Scene Graphs with LLM}
Given a set of region-specific narratives and a global narrative
, we can follow the previous pipeline (see \cref{fig:pipeline})
to extract relationship triplets using a scene parser. 
However, this method will meet with troubles.
Take the example in \cref{fig:gpt4sgg}, \textit{a cake made of books} can be parsed as 
(\textit{cake}, \textit{made of}, \textit{book}) but there is no
\textit{cake} in the image and there are several \textit{books}.
Another challenge is to keep the semantic consistency, \ie, an unique instance can be captioned in different ways (\eg, ``\textit{man}'', ``\textit{people}'', ``\textit{person}''
for \textit{person.2} in \cref{fig:gpt4sgg}
) but the model should correctly associate the alias.

To perform relation reasoning with fewer misjudgments and semantic consistency, we utilize an LLM to synthesize scene graphs from those narratives and objects' information. 
Specifically, we construct a prompt template shown in Tab. \ref{tab:prompt}, which requires the model to use information about objects (such as their categories and bounding boxes) and both holistic and region-specific descriptions to establish relationships between objects. 
To distinguish multiple instances, we encode the object information as ``[\textit{cateogry}].[\textit{number}]: [\textit{box}]'' in the prompt. 
This unique representation enables the LLM to correctly match visual regions with relationship triplet, especially when multiple instances with the same category appear in an image.

The LLM is also required to maintain logical consistency to circumvent impossible or nonsensical relationships.  
With this carefully designed prompt, GPT-4 can synthesize scene graphs based on the provided image information.  
An example of this process is shown in Tab. \ref{tab:examples}. 
The example shows that GPT-4 generates a comprehensive and accurate scene graph, and provides a reasonable explanation for inferring latent relationships not explicitly mentioned in the captions.
\begin{table}[t]
\centering
\definecolor{colorbg@intro}{RGB}{240,240,240}     
\caption{Prompt for synthesizing scene graphs.
For clarity, we only show the main part of the prompt.
A complete prompt template can be found in supplementary materials. 
}
\begin{tikzpicture}[
    node distance=1cm and 0.8cm,
    auto, 
    description/.style={
        rectangle, rounded corners=2pt, draw, thick, fill=colorbg@intro, 
        text width=0.95\textwidth, 
        align=left, font=\footnotesize,
    },
] 
\node[description] {  
\textcolor{blue}{messages} = [\{
    \verb|"|\textbf{role}\verb|"|: \verb|"|system\verb|"|,
    \verb|"|\textbf{content}\verb|"|: \verb|"|You are a helpful AI visual assistant. Now, you are seeing image data. Each image provides a set of objects, and a set of captions for global and localized descriptions.\verb|"|
  \}, \\ 
    \smallskip\{ \verb|"|\textbf{role}\verb|"|: \verb|"|user\verb|"|, \verb|"|\textbf{content}\verb|"|:  f\verb|"""|Extract relationship triplets from image data, each characterized by a unique ``image\_id", image dimensions,  a set of objects consisting of categories (formatted as ``[category].[number]") and bounding boxes (in ``xyxy" format). Each image data includes $\cdots$
    
Here are the requirements for the task:
1. Process each image individually: $\cdots$
2. Infer interactions and spatial relationships: $\cdots$
3. Maintain logical consistency: $\cdots$
5. Output should be formatted as a list of dicts in JSON format, containing ``image\_id" and ``relationships" for each image.

Example output: \{\textcolor{blue}{example\_output}\} 

Ensure that each image's data is processed and outputted separately to maintain clarity and accuracy in the relationship analysis.

\verb|###| Input:
``
\{\textcolor{blue}{Input}\} ''
\verb|###| Output:
    \} \verb|"""|
    ]
\\ 

};
\end{tikzpicture}
\label{tab:prompt}
\end{table}
 
\input{tables/gpt4sgg_examples}
\subsection{Training SGG Models}
After obtaining the generated scene graphs by an LLM, any standard SGG models can be trained with localized objects and relationships. 
A prerequisite for SGG models under a closed-set setting is that 
the localized objects and generated relationships should be mapped into
target categories before training. 
\subsection{Instruction Tuning Private LLMs}
We use Low-Rank Adaptation (LoRA) \cite{lora22} to fine-tune a private and local LLM, Llama 2 \cite{touvron2023llama},  with the instruction-following data generated by GPT-4.
This process involves adapting Llama 2 to better align with specific instruction sets, enhancing its ability to synthesize scene graphs from textual image data. 
\section{Experiments}
\subsection{Datasets}
\noindent\textbf{VG150} \cite{xu2017scene} is the widely used dataset for evaluating SGG models, which consists of $108, 777$ images with $150$ object categories and $50$ predicate categories. 
For standard split, VG150 utilizes $70\%$ of its images for training, $5,000$ for validation, and the remainder for testing.
VG150 contains ${\sim}257k$ instances of relationship triplets, covering spatial relationships and interactions.

\noindent\textbf{COCO Captions} \cite{chen2015microsoft} consists of ${\sim}117k$ images, in which each image is equipped with five manual captions. Previous works \cite{zhang2023learning,zhong2021learning} utilize a scene parser \cite{mao2018parser} to extract relation triplets from the caption data, yielding ${\sim}181k$ instances of relationship triplets with ${\sim}44k$ phrases and ${\sim}2.5 k$ relations. 
For clarity, we refer to this data as ``\textbf{COCO@Parser} \cite{mao2018parser}''.

\noindent\textbf{COCO@GPT} is derived from COCO with annotated bounding boxes containing ${\sim}94k$ images. 
The model card for generating COCO@GPT is \textit{GPT-4 Turbo}.
The COCO@GPT includes ${\sim}394k$ instances of relationship triplets with $80$ object categories and  ${\sim}4.7k$ predicate categories.

\noindent\textbf{VG@GPT} is derived from VG150 using GPT4SGG, containing 
${\sim}47k$ images.  
This dataset includes ${\sim}227k$ instances of relationship triplets with
$150$ object categories and ${\sim}2.4k$ predicate categories.

 We provide the statistics of head-10 / tail-10 categories of the VG150 dataset, compared across VG@GPT, 
COCO@Parser, and COCO@GPT data in \textbf{supplementary materials}. 
It can be found that COCO@Parser is biased to conjunction words such as  ``with'', ``in'', ``from'',  while less focus on interactions like ``holding'', ``wearing'', \textit{etc.} 
This defect makes it hard for the SGG model to learn complex relationships. Conversely, COCO@GPT provides rich instances for learning such complex relationships.
\subsection{Experimental Setup}
\textbf{Model settings}. we mainly compare two SGG models with \textit{GPT4SGG}:
1) $\text{VS}^3$ \cite{zhang2023learning} extends the object detector of the SGG model from closed-set to open-vocabulary.
2) OvSGTR \cite{chen2023expanding} advances the SGG from closed-set to both object and relation-aware open-vocabulary setting.
For a fair comparison, we follow the official implementations to conduct experiments.

\noindent\textbf{Metric.} 
We use the protocol \textbf{SGDET} \cite{xu2017scene, tang2020unbiased} 
(also known as \textbf{SGGen}) 
to measure the capability of SGG models. 
Recall@K (R@K, K=20/50/100) and mean Recall@K (mR@K, K=20/50/100) are reported. 
mR@K reflects the capability of SGG models to address long-tail bias.

\subsection{Experimental Results}
\begin{table}[t]
    \centering
    \caption{Comparison with state-of-the-art methods on VG150 test set. \ding{71} marks fully supervised methods. } 
   \resizebox{0.9\textwidth}{!}
    {
    \begin{tabular}{l|c|c|ccc|ccc}
    \toprule
         SGG model&  Training Data & Grounding & 
         \multicolumn{3}{c|}{R@20/50/100} & 
         \multicolumn{3}{c}{mR@20/50/100} \\
    \midrule
         LSWS  \cite{yelinguistic}& \multirow{6}{*}{COCO Caption}   & - & - & 3.28 & 3.69   
         &  -   &  -  & -\\ 
         MOTIFS \cite{zellers2018neural}&   & Li \etal  \cite{li2022integrating} & 5.02 & 6.40 & 7.33   & -   &  -  & -\\ 
         Uniter \cite{chen2020uniter}&   & SGNLS \cite{zhong2021learning} & - & 5.80 & 6.70  &  -   &  -  & -\\ 
         Uniter \cite{chen2020uniter} & 
             & Li \etal  \cite{li2022integrating} & 5.42 & 6.74 & 7.62  &   -   &  -  & -\\ 
         $\text{VS}^3_{\text{(Swin-L)}}$  \cite{zhang2023learning}  &   & GLIP-L \cite{li2022grounded}  & 5.26 & 6.70 & 7.91  & 1.97  &  2.38  & 2.70\\ 
        \multirow{2}{*}{$\text{OvSGTR}_\text{(Swin-B)}$ \cite{chen2023expanding} } &  & Grounding--   &  \multirow{2}{*}{6.85}
          &   \multirow{2}{*}{9.33}  & \multirow{2}{*}{11.47}   
          &  \multirow{2}{*}{1.28}   &  \multirow{2}{*}{1.79}
          & \multirow{2}{*}{2.18} \\ 
        &  & {DINO}\cite{liu2023grounding}  &    
          &      &       &     &   &  \\  
       \hline 
       \rowcolor{lightgray}
       $\text{VS}^3_{\text{(Swin-L)}}$  \cite{zhang2023learning}    &  & 
       &    \textbf{5.07} &   \textbf{7.40}   &   \textbf{9.50} &   \textbf{1.30} & \textbf{1.93} &  \textbf{2.42} \\ 
        \rowcolor{lightgray}
        $\text{OvSGTR}_\text{(Swin-B)}$ \cite{chen2023expanding}  & \multirow{-2}{*}{COCO@GPT}    &\multirow{-2}{*}{-} 
           &  ${\textbf{7.65}}$  & ${\textbf{10.10}}$    &  ${\textbf{11.73}}$    &   ${\textbf{2.92}}$ &   ${\textbf{3.84}}$  & ${\textbf{4.69}}$ \\
          \hhline{=========} 
        IMP \cite{xu2017scene}\ding{71}&   &\multirow{4}{*}{-}  &  17.73  & 25.48  & 30.71 & 2.66 & 4.10 & 5.29  \\  
        $\text{VS}^3_{\text{(Swin-L)}}$  \cite{zhang2023learning} \ding{71}  & VG150   & &
        27.34 & 36.04 & 40.88 &  4.43 & 6.45 & 7.81 \\      
        $\text{OvSGTR}_\text{(Swin-B)}$ \cite{chen2023expanding} \ding{71} & (${\sim}76k$ ) & &
         27.80 & 36.40 &  42.40 & 5.24 & 7.41 & 8.98 \\ 
        \midrule 
         LSWS  \cite{yelinguistic} &   & - & - & 3.85 & 4.04 & - & - & - \\ 
        MOTIFS \cite{zellers2018neural} &   & SGNLS \cite{zhong2021learning} &  
        6.31 & 8.05 & 9.21 & - & - & - \\ 
        MOTIFS \cite{zellers2018neural} &  &
        Li \etal  \cite{li2022integrating} 
        & 8.25 & 10.50 & 11.98 & - & - & - \\ 
        Uniter \cite{chen2020uniter} & VG Caption &
        SGNLS \cite{zhong2021learning} & 
        - & 9.20  & 10.30 & - & - & - \\ 
        Uniter \cite{chen2020uniter}  
        & (${\sim}73k$)  & Li \etal  \cite{li2022integrating} & 8.90 & 10.93 & 12.14 & - & - & - \\ 
        
        $\text{VS}^3_{\text{(Swin-L)}}$  \cite{zhang2023learning}
        &  & GLIP-L \cite{li2022grounded}
        &11.31 &16.00 &19.85& 2.39 & 3.80 & 4.87\\ 
        $\text{OvSGTR}_\text{(Swin-B)}$ \cite{chen2023expanding} &  
         &  GLIP-L \cite{li2022grounded}   &  16.36 & 22.14 & 26.20 & 3.80 &5.24 & 6.25   \\   
        \hline 
        \rowcolor{lightgray}
       $\text{VS}^3_{\text{(Swin-L)}}$  \cite{zhang2023learning}
        & VG@GPT&   
        &\textbf{17.77}   & \textbf{22.42}  &  \textbf{25.29}
        & \textbf{4.24}  & \textbf{5.82}  &  \textbf{6.97} \\   
        \rowcolor{lightgray}
         $\text{OvSGTR}_\text{(Swin-B)}$ \cite{chen2023expanding} & (${\sim}46k$)  & \multirow{-2}{*}{-}
         & \textbf{20.12}  & \textbf{25.03}  & \textbf{28.84} & \textbf{5.68} & \textbf{7.14} &  \textbf{8.22} \\   
   \bottomrule
    \end{tabular}
    }
    \label{tab:lang}
\end{table}

\begin{figure}[t]
\definecolor{cbBlue}{RGB}{0,107,164} 
\definecolor{cbRed}{RGB}{200,82,0} 
\definecolor{cbGreen}{RGB}{16,115,0} 
\centering
\resizebox{0.98\textwidth}{!}
{
\begin{tikzpicture}[
    node distance=1.5cm and 1cm,
    auto, 
    arrow/.style={thick,->},
    filled node/.style={circle, draw, thick, fill=colorfillednode, scale=0.6},
    llmnode/.style={rectangle, rounded corners=2pt, draw, thick, fill=colorllm, text width=2.5cm, align=center, font=\Large, drop shadow},
    sggnode/.style={rectangle, rounded corners=2pt, draw, thick, fill=colorsgg, text width=2.5cm, align=center, font=\Large, drop shadow},
    description/.style={rectangle, rounded corners=2pt, draw, thick, fill=colorbg, text width=3.5cm, align=left, font=\footnotesize}
]
\definecolor{colorllm}{RGB}{79,129,189}     
\definecolor{colorsgg}{RGB}{155,187,89}     
\definecolor{colorarrow}{RGB}{192,80,77}    
\definecolor{colorbg}{RGB}{217,217,217}     
\definecolor{colorfillednode}{RGB}{128, 100, 162} 

\definecolor{colorbbox}{RGB}{255, 230, 153} 

\definecolor{colorArrow}{RGB}{64,64,64} 
\definecolor{color_flow}{RGB}{173, 216, 230}

\tikzset{
  myarrow/.style={
    ->,                    
   -{Latex[round]},             
    thin, 
    draw=colorArrow,       
    line width=1.5pt,      
    text=black,            
    shorten >=0.4pt,         
    shorten <=0pt,           
  }
}

\node (n0)  {
    \begin{tikzpicture}[yscale=0.9]
    \begin{axis}[
        width=0.3\textwidth,
        height=1.1in,
        scale only axis,
        clip=false,
        separate axis lines,
        axis on top,
        xmin=0.7,
        xmax=3.3,
        xtick={1,2,3},
        x tick style={draw=none},
        xticklabels={GPT4SGG, w. WordNet, w. lookup},
        xticklabel style={font=\tiny\bfseries, align=center, scale=0.8}, 
        ylabel style={font=\scriptsize},
        ylabel shift=-2pt, 
        ymin=11,
        ymax=12.3,
        ylabel={Recall (\%)},
        ylabel style={font=\tiny\bfseries},
        nodes near coords,
        nodes near coords align={vertical},
        enlarge x limits=0.1,
        grid style=dashed,
        every axis plot/.append style={
          ybar,
          bar width=10pt,
          bar shift=0pt,
          fill
        }
      ]
      \addplot[cbGreen]coordinates {(1, 11.43)};
      \addplot[cbBlue]coordinates{(2, 11.70)};
      \addplot[cbRed]coordinates{(3, 12.09)};
    \end{axis}
    \end{tikzpicture}
};
  \node [below=of n0, yshift=16mm, xshift=5mm] {(a)};
  
    \node (n1) [right =of n0, xshift=-5mm, yshift=3mm] {
        \includegraphics[width=0.2\textwidth]{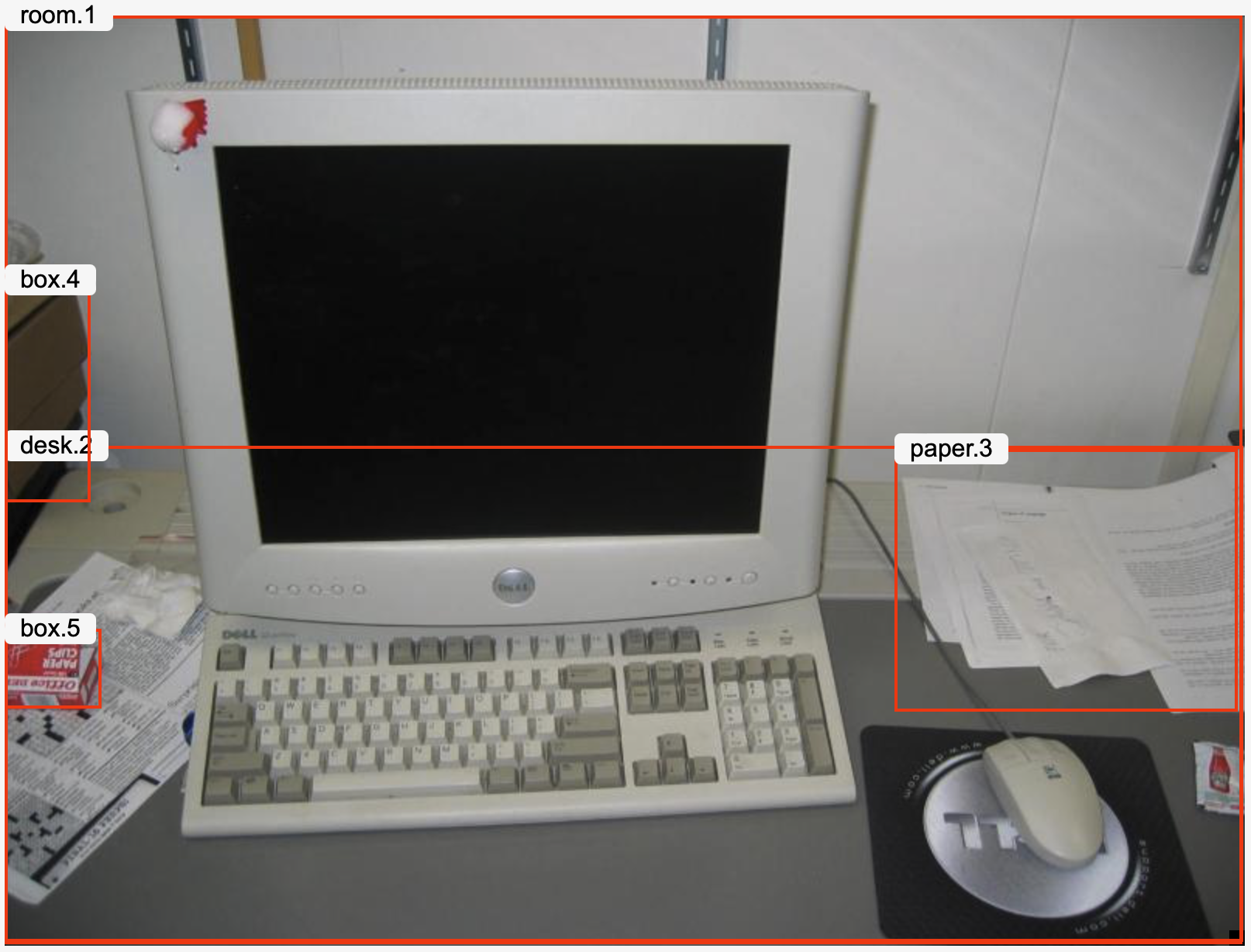}
    };

\node[filled node, right of=n1, xshift=15mm, yshift=10mm] (desk2) {};
\node [above=of desk2, yshift=-14.5mm] {\tiny desk.2}; 

\node[filled node, right of=desk2, xshift=5mm] (paper3) {};
\node (paper3text) [above=of paper3, yshift=-14.5mm] {\tiny  paper.3};

\node[filled node, below  of=desk2, yshift=-5mm] (box4) {};
\node (box4text) [below=of box4, yshift=15mm] {\tiny  box.4}; 

\node[filled node, below of=paper3, yshift=-5mm] (box5) {};
\node (box5text) [below=of box5, yshift=15mm] {\tiny box.5}; 

\draw [->, thick, >=stealth] (desk2) -- node[midway, sloped, above, yshift=-0.5mm] {\tiny supporting} (paper3);
\draw [->, thick, >=stealth] (desk2) -- node[midway, sloped, above] {\tiny supporting} (box4);
\draw [->, thick, >=stealth] (desk2) -- node[midway, sloped, above] {\tiny supporting} (box5);

\node[filled node, right of=box5, xshift=4mm] (desk2b) {};
\node (d2btext) [below=of desk2b, yshift=15mm] {\tiny desk.2}; 

\node[filled node, right of=paper3,  xshift=4mm] (paper3b) {};
\node (p3btext) [above=of paper3b, yshift=-14.5mm] {\tiny  paper.3}; 

\draw [->, thick, >=stealth] (paper3b) -- node[midway, sloped, above, rotate=90, yshift=-2mm] {\tiny on/above} (desk2b);

\node (cap1) [below of=n1, yshift=2.5mm] {\scriptsize Image }; 
\node (cap2) [right of=cap1, xshift=10mm] {\scriptsize GPT4SGG}; 
\node (cap3) [right of =cap2, xshift=1.5mm] {\scriptsize GT }; 

\node [below of=cap2, yshift=7.8mm, xshift=-7mm] {(b)};

\begin{scope} [on background layer]
    \node (bg3) [draw, dashed,  rounded corners,  
    fit=(n1)(paper3text)(box4text)(box5text)(p3btext)(d2btext)(cap1)(cap2)(cap3),  inner sep=2pt,   
    xshift=0mm, yshift=0.5mm, fill=colorbg, 
    ] {};
\end{scope}

\begin{scope} [on background layer]
    \node (bg1) [draw,  rounded corners,  fit=(paper3text)(box4text)(box5text),  inner sep=0pt,   
    xshift=0mm, yshift=0.1mm, color=red, 
    ] {};
\end{scope}

\begin{scope} [on background layer]
    \node (bg2) [draw,   rounded corners,  fit=(p3btext)(d2btext),  inner sep=0pt,   
    xshift=0mm, yshift=0.1mm, color=blue!80, 
    ] {};
\end{scope}

\end{tikzpicture}
}
\caption{Comparison with manual annotation on VG150 validation set.
(a) quantitative results of using different semantic matching strategies;
(b) an example of low recall rate based on manual annotation. 
}
\label{fig:recall_comparison}
\end{figure}
\textbf{Quantitative results.} 
\cref{tab:lang} reports a comparison with state-of-the-art models on VG150 test set. 
From the result, \textit{GPT4SGG} significantly improves the performance of SGG models. 
For instance,  OvSGTR \cite{chen2023expanding} (Swin-B) with few images outperforms itself trained on VG caption (25.03/28.84 vs. 
22.14/26.20, R@50/100).  
The improvement of mR@K also indicates that \textit{GPT4SGG} 
alleviates the long-tail problem. 
This improvement is attributed to our novel approach can generate more accurate and comprehensive scene graphs. 

Another observation is that the performance gain on COCO is less significant than on the VG150 dataset. 
The reason is twofold. First,  COCO only has 80 categories for \textit{GPT4SGG} and only 66 categories can be matched into VG150, 
resulting in worse recall of objects. 
Another reason is the distribution discrepancy between the two datasets, which is well-known from previous methods.

\noindent\textbf{Qualitative results.} We provide samples generated by \textit{GPT4SGG} and corresponding results generated by Scene Parser \cite{mao2018parser} and GPT-4 \cite{mao2018parser}, as shown in Fig. \ref{fig:eg_comp}. From these samples, \textit{GPT4SGG} can generate more accurate and complete scene graphs. 
The qualitative comparison also presents the ambiguity issue in grounding objects and the relationship density difference between ours and the previous pipeline.
\subsection{Ablation Studies} 
\noindent\textbf{How close of GPT4SGG compared to manual annotation?}
To quantitatively verify the quality of generated data by \textit{GPT4SGG},  we make a comparison on the VG150 validation set. 
\cref{fig:recall_comparison} (a) reports the average recall rate of manual annotation.
To calculate the recall based on semantic meaning, we need to align synonyms and antonyms of relationships. 
For example,  ``\textit{next to}'' (not in VG150) and ``\textit{near}'' (in VG150) 
are synonyms; ``\textit{ridden by}'' (not in VG150) and ``\textit{riding}''(in VG150) are antonyms. 
In \cref{fig:recall_comparison} (a), \textit{w. WordNet} utilizes WordNet \cite{miller1995wordnet} synsets matching and \textit{w. lookup} builds a lookup table by leveraging GPT-4
(please see \textbf{ supplementary materials}). 
From the result,  \textit{w. lookup} works better than \textit{w. WordNet} (12.09 vs. 11.70); as there are many complicate relationships, \eg,  ``\textit{standing next to}'' can be mapped into ``\textit{near}''.

Moreover, \emph{low recall rate does not mean low-quality scene graphs}. 
As shown in \cref{fig:recall_comparison} (b), 
for the same scene,  the prediction selects ``\textit{supporting}'' to describe the relationship between ``\textit{desk.2}'' and ``\textit{paper.3}'' 
while GT (ground truth) adopts ``\textit{on} / \textit{above}'' to describe it. Both are correct but present different preferences.

\noindent \textbf{Impact of Ambiguity in grounding.}
To showcase the impact of ambiguity in grounding, we select ${\sim}27k$ images from the VG150 training set containing multiple instances of the same category.
These images are referred to as ``VG@Poly''.
This selection process ensures we focus on scenarios where ambiguity in visual-language alignment is prevalent.
\cref{tab:ablation_amb} reports the performance of SGG models training on VG@Poly.
The performance drop (Row 2 vs. Row 1) highlights the limitations of the previous pipeline in generating accurate scene graphs under conditions of grounding ambiguity. 
In contrast, \textit{GPT4SGG} demonstrates a significant improvement, effectively mitigating this challenge. 
For example,  both $\text{VS}^3$ and OvSGTR achieve comparable or even better performance on mR@K compared to fully supervised training (Row 1).

Another interesting finding is that the performance of models trained on only 27k images is close to that of a large-scale dataset (46k images). 
This minor difference indicates ``VG@Poly'' (365k bounding boxes, 151k triplets) has higher density in terms of nodes and edges, compared to other images in the VG150 training set 
(Of its 46k images, there are 549k bounding boxes and 225k triplets). 

\begin{table}[t]
    \centering
    \caption{Ablation study of the effect of ambiguity in grounding. 
    All models are trained on VG@Poly (${\sim}27k$ images) and tested on VG150 test set. 
    Compared to manual annotation, \textit{GPT4SGG} obtains a recall rate $13.4\%$ on VG@Poly.  
    }    
   \resizebox{0.9\textwidth}{!}
    {
    \begin{tabular}{l|c|c|ccc|ccc}
    \toprule
         SGG model& Supervision&\#Triplets & 
          \multicolumn{3}{c|}{R@20/50/100}  &
          \multicolumn{3}{c}{mR@20/50/100}   \\
    \midrule
        $\text{VS}^3_{\text{(Swin-L)}}$  \cite{zhang2023learning}
        &   \multirow{2}{*}{Annotation}& \multirow{2}{*}{157k}   & 24.62 & 32.93   & 38.07   & 3.67
          & 5.50  &  6.84 \\ 
         $\text{OvSGTR}_\text{(Swin-B)}$ \cite{chen2023expanding}  &
         &       &27.46   &  35.98 & 41.37
         & 5.04  &  6.78 & 7.97  \\  \hline 
       $\text{VS}^3_{\text{(Swin-L)}}$  \cite{zhang2023learning}
        &  Parser \cite{mao2018parser}+& \multirow{2}{*}{386k}&   
        10.63 &  14.97 & 18.83  & 
         1.65& 2.63 &   3.44\\ 
  $\text{OvSGTR}_\text{(Swin-B)}$ \cite{chen2023expanding} &GLIP-L \cite{li2022grounded} & &  15.63  &  20.58  & 24.20  &  3.33 & 4.48
         &  5.38   \\ \hline 
        \rowcolor{lightgray}
        $\text{VS}^3_{\text{(Swin-L)}}$  \cite{zhang2023learning}&   
        &   
        & \textbf{17.74} & \textbf{22.27}  & \textbf{25.17}  & 
        \textbf{4.07} & \textbf{5.38} & \textbf{6.48}  \\  
         \rowcolor{lightgray}
     $\text{OvSGTR}_\text{(Swin-B)}$ \cite{chen2023expanding} 
       & \multirow{-2}{*}{\textit{GPT4SGG}}     & \multirow{-2}{*}{151k}&  
       \textbf{19.19} &  \textbf{23.96} & \textbf{27.51}
         & \textbf{5.32} & \textbf{6.77}  & \textbf{7.80} \\
   \bottomrule
    \end{tabular}
    }
    \label{tab:ablation_amb}
\end{table}
\begin{figure}[t]
\definecolor{cbBlue}{RGB}{0,107,164} 
\definecolor{cbRed}{RGB}{200,82,0} 
\definecolor{cbGreen}{RGB}{16,115,0} 
\definecolor{histColor}{RGB}{187, 85, 102}
\centering
\resizebox{\textwidth}{!}{
\begin{minipage}{0.44\textwidth}
\centering
\begin{tikzpicture}[xscale=0.6, yscale=0.5] 
    \begin{axis}[
        xlabel={Maximum number of RoIs},
        ylabel={Recall (\%)},
        ylabel style={font=\bfseries},
        xlabel style={font=\bfseries},
        xmin=2, xmax=35,
        ymin=2, ymax=13, 
        xtick={5,10, 20, 30},
        ytick={2,4,6,8,10,12},
        legend pos=north west,
        axis y line*=left,
        legend style={
            at={(0, 1.31)},
            font=\scriptsize\bfseries, align=center}, 
        legend cell align={left},
    ]

    \addplot[
        color=cbGreen,
        mark=diamond,
        thick,
        ] coordinates {
        (5,6.38)(10,6.74)(20,7.80) (30, 9.14)
    };
    \label{plot_recall}
    \addlegendentry{Recall (GPT-3.5+BLIP-2/2.7b)}

    \node at (axis cs:5,6.38) [above] {6.38};
    \node at (axis cs:10,6.74) [above] {6.74};
    \node at (axis cs:20,7.80) [below, xshift=5mm] {7.80};
    \node at (axis cs:30, 9.14) [below] {9.14}; 

    \addlegendimage{/pgfplots/refstyle=plot_tokens}\addlegendentry{Prompt Tokens}

    \addplot[
        color=red,
        mark=*,
        thick,
        ] coordinates {
        (20, 12.09)
    };
    \addlegendentry{Recall (GPT-4)}

    \node at (axis cs:20,12.09) [above] {12.09};

    \addplot[
        color=blue!50,
        mark=diamond,
        thick,
        ] coordinates {
        (20, 7.27)
    };
    \addlegendentry{Recall (GPT-3.5+GIT-Base)}

    \node at(axis cs:20, 7.27)[below] {7.27};
    \end{axis}

    \begin{axis}[
        axis y line*=right,
        axis x line=none,
        ylabel={Prompt Tokens (M)},
        ylabel style={font=\bfseries},
        xmin=2, xmax=35,
        ymin=2, ymax=6,
        ytick={2,3,4,5},
        legend pos=north east,
    ]

    \addplot[
        color=cbBlue,
        mark=triangle*,
        dashed
        ] coordinates {
        (5,2.393)(10,3.222)(20,4.450) (30, 5.279)
    };
    \label{plot_tokens}

    \node at (axis cs:5,2.393) [below] {2.39};
    \node at (axis cs:10,3.222) [below] {3.22};
    \node at (axis cs:20,4.450) [below] {4.45};
    \node at (axis cs:30, 5.279) [above] {5.28};
    
    \end{axis}
\end{tikzpicture}
\subcaption{recall rate under different settings.}
\end{minipage}  
    \begin{minipage}{0.55\textwidth} 
    \centering
    \vspace{1.95em}
        \begin{tikzpicture}[xscale=0.7, yscale=0.62]
            \begin{axis}[
                ybar, 
                bar width=1pt, 
                xlabel={\#edges},
                ylabel={Frequency},
                xlabel style={font=\bfseries},
                ylabel style={font=\bfseries},
                width=\textwidth, 
                height=6cm, 
                xtick={0,10,20,30,40,50}, 
                xticklabels={0,10,20,30,40,50}, 
                ymin=0, ymax=900, 
                enlarge x limits=0.05, 
                title style={font=\bfseries},
                ]
\addplot [color=histColor] coordinates {
    (1,818) (2,622) (3,559) (4,462) (5,383) (6,323) (7,263) (8,233) (9,199) (10,188)
    (11,154) (12,118) (13,115) (14,102) (15,73) (16,57) (17,54) (18,39) (19,34) (20,28)
    (21,25) (22,14) (23,17) (24,10) (25,12) (26,13) (27,14) (28,6) (29,8) (30,7)
    (31,5) (32,3) (33,3) (34,4) (35,3) (36,1) (37,4) (38,4) (39,2) (40,0)
    (41,2) (42,1) (43,1) (44,1) (45,1) (46,0) (47,0) (48,3)
};
            \end{axis}
        \end{tikzpicture}
        \subcaption{histogram of \#edges of VG150 val. set.}
    \end{minipage}
    }
    \caption{Impact of Caption. Considering the cost of ChatGPT APIs, we choose \textit{GPT-3.5-turbo} as the main LLM to probe the performance under different settings. 
    }
    \label{fig:caption}
\end{figure}
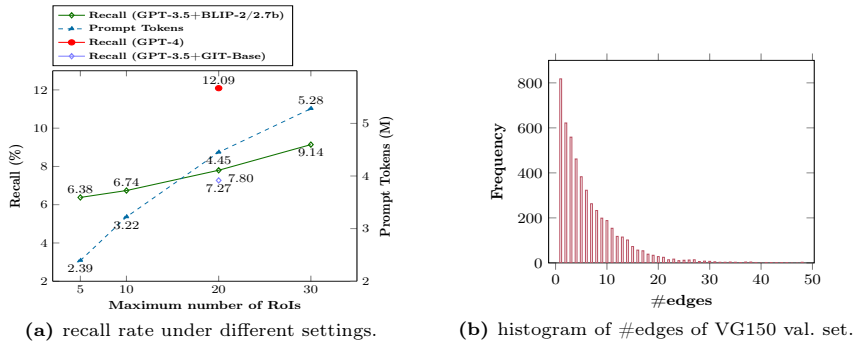
\input{pics/eg_comp}

\noindent\textbf{Impact of Caption.} 
\cref{fig:caption} reports the result of using different captioning models and 
maximum number of region descriptions. 
Without surprise, increasing the maximum number of regions (to cover the number of ground-truth edges) obtains better recall performance. 
Considering the trade-off between prompt token usage and performance, we choose a maximum number N=20 to generate data by GPT-4.
We also compare two captioning models: BLIP-2 / opt-2.7b \cite{blip2} and GIT-Base \cite{wang2022git} .
With GPT-3.5,  using BLIP-2 can obtain a better result than using GIT-Base 
(7.80 vs. 7.27). 
The difference indicates that the captioning model in \textit{GPT4SGG} plays an important role to obtain accurate scene graphs.

\noindent\textbf{Source of LLM.} 
This study compares the capability of different LLMs, including GPT-3.5 \cite{openai2023gpt35}, GPT-4 \cite{gpt4}, and Llama 2 \cite{touvron2023llama} to synthesize scene graphs.
Without fine-tuning, \textit{GPT-3.5 turbo} achieves $7.78\%$ recall on VG150 validation set (4,808 images) with 4,537 valid responses;
Llama 2 - 13B \cite{touvron2023llama} obtains  recall rate $8.02\%$ but with only $3,876$ valid responses;
\textit{GPT-4 turbo} achieves the highest recall $12.09\%$ with $4,760$ valid responses.
The result shows Llama 2 has the worst instruction-following capability among the three LLMs. 

We fine-tuned Llama 2 - 13B \cite{touvron2023llama} using instruction-following data generated by GPT-4,
resulting in $12.06\%$ recall performance on the VG150 validation set. From the result,  the instruction-tuned Llama 2 \cite{touvron2023llama} can achieve comparable performance to GPT-4 \cite{gpt4}, providing a cost-effective and private option for the LLM in \textit{GPT4SGG}.

\noindent \textbf{Impact of Object detection/grounding.}
We compare the quality of generated data between manual annotation and object detectors.
For object detectors, we choose one state-of-the-art model,
Grounding DINO \cite{liu2023grounding} (\textit{w.} Swin-B), which obtains 22.2 AP50 on VG150 validation set without fine-tuning. 
Compared to using manual annotation, 
\textit{GPT4SGG} (\textit{w.} GPT-3.5-turbo) using the detectors' prediction only obtains a recall of 3.11, far from the baseline of 7.80.
The recall drop mainly lies in the low recall rate of objects 
(the detector only has a $56.8\%$ average recall at IoU threshold 0.5). 
Besides, as discussed before,  a low recall rate compared to manual annotation does not mean low-quality scene graphs.

\subsection{Discussion}
This work introduces a novel pipeline to learning scene graphs from natural language descriptions. 
Despite its advantages,  there remain some limitations. 
First, we do not consider using multi-modal LLMs like GPT-4V to process images directly. 
Second, the performance of such a pipeline is highly affected by the ability of LLMs.
Finally, due to the bias of manual annotation and machine generation, 
the current evaluation protocol only using recall cannot comprehensively evaluate the quality of generated scene graphs.

\noindent\textbf{Future work.} Limited by computation resources, 
constructing large-scale SGG datasets using the open-sourced Llama and existing large-scale object detection datasets like Open Images \cite{OpenImages}, LVIS \cite{gupta2019lvis} has yet to be explored. 
We believe such large-scale SGG pre-training scheme can boost the generalization ability of SGG models. 

\section{Conclusion}
 In this work, we propose a simple yet effective framework, \textit{GPT4SGG}, 
to synthesize scene graphs from textual representation for images.
This framework decomposes a complex scene of an image into 
a set of simple regions to obtain their descriptions.
With these region-specific narratives, a global narrative, and object information, an LLM is utilized to conduct a relationship 
reasoning and synthesizing the final scene graph. 
Our framework effectively addressed the ambiguity issue in
grounding unlocalized objects and alleviated the long-tail bias.
Experimental results demonstrate \textit{GPT4SGG} can generate more accurate and comprehensive scene graphs. 

\clearpage 
\appendix

\section{Details of Methods}
\subsection{Prompt for LLMs}
We provide the prompt template in \cref{tab:prompt}.
This prompt is designed by human experience and optimized with conversations between 
human and ChatGPT (GPT-4, December 2023). 

\subsection{Example of Synthesizing Scene Graphs}
A complete example used in this work can refer to \cref{tab:examples}.

\begin{table}[t]
    \centering
    \caption{Comparison with state-of-the-art methods on VG150 test set. \ding{71} marks fully supervised methods. } 
   \resizebox{0.9\textwidth}{!}
    {
    \begin{tabular}{l|c|c|ccc|ccc}
    \toprule
         SGG model&  Training Data & Grounding & 
         \multicolumn{3}{c|}{R@20/50/100} & 
         \multicolumn{3}{c}{mR@20/50/100} \\
    \midrule
          IMP \cite{xu2017scene}\ding{71}&  VG150  &\multirow{1}{*}{-}  &  17.73  & 25.48  & 30.71 & 2.66 & 4.10 & 5.29  \\  
           IMP \cite{xu2017scene} &  VG Caption  &GLIP-L \cite{li2022grounded}&    6.82 &  9.79  &12.12  &  0.98 &  1.55 &2.12   \\  
           IMP \cite{xu2017scene} &  VG@GPT &\multirow{1}{*}{-}  &  12.46   & 16.24   & 18.99  &   \textbf{3.97}&  \textbf{5.01} &  \textbf{5.82}  \\  
           \bottomrule 
    \end{tabular}
    \label{tab:exp}
}
\end{table}

\begin{table}[t]
\centering
\definecolor{colorbg@intro}{RGB}{240,240,240}     
\caption{Prompt for synthesizing scene graphs. 
We only provide an output example in the instruction, while the input example is excluded for no bias introduction.
}
\begin{tikzpicture}[
    node distance=1cm and 0.5cm,
    auto, 
    description/.style={
        rectangle, rounded corners=2pt, draw, thick, fill=colorbg@intro, 
        text width=\textwidth, 
        align=left, 
        scale=1.0
    },
] 
\node[description] {  
\textcolor{blue}{messages} = [\{
    \verb|"|\textbf{role}\verb|"|: \verb|"|system\verb|"|,
    \verb|"|\textbf{content}\verb|"|: \verb|"|You are a helpful AI visual assistant. Now, you are seeing image data. Each image provides a set of objects, and a set of captions for global and localized descriptions.\verb|"|
  \}, \\ 
    \smallskip\{ \verb|"|\textbf{role}\verb|"|: \verb|"|user\verb|"|, \verb|"|\textbf{content}\verb|"|:  f\verb|"""|Extract relationship triplets from image data, each characterized by a unique ``image\_id", image dimensions,  a set of objects consisting of categories (formatted as ``[category].[number]") and bounding boxes (in ``xyxy" format). Each image data includes a global description for the entire image and localized descriptions for specific regions (notated as ``Union(name1:box1, name2:box2)", keys with ``;" in captions like ``Union(name1:box1, name2:box2); Union(name3:box3, name4:box4)" refer to multiple union regions share the same caption). 
    
Here are the requirements for the task:
1. Process each image individually: Focus on one image at a time and give a comprehensive output for that specific image before moving to the next.
2. Infer interactions and spatial relationships: Utilize objects' information and both global and localized descriptions to determine relationships between objects(e.g., ``next to'', ``holding", ``held by", etc.).
3. Maintain logical consistency: Avoid impossible or nonsensical relationships (e.g., a person cannot be riding two different objects simultaneously, a tie cannot be worn by two persons, etc.). 
4. Eliminate duplicate entries: Each triplet in the output must be unique and non-repetitive.
5. Output should be formatted as a list of dicts in JSON format, containing ``image\_id" and ``relationships" for each image.

Example output:
`
[
{``image\_id": ``123456",
 ``relationships": [
    {``source": ``person.1", ``target": ``skateboard.2", ``relation": ``riding"},
    {``source": ``person.4", ``target": ``shirt.3", ``relation": ``wearing"},
    {``source": ``person.2", ``target": ``bottle.5", ``relation": ``holding"},
    {``source": ``person.4", ``target": ``bus.1", ``relation": ``near"},
  ]
},
{``image\_id": ``23455",
 ``relationships": [
    {``source": ``man.1", ``target": ``car.1", ``relation": ``driving"}
  ]
}
]
'

Ensure that each image's data is processed and outputted separately to maintain clarity and accuracy in the relationship analysis.

\verb|###| Input:
``
\{\textcolor{blue}{Input}\} ''
\verb|###| Output:
    \} \verb|"""|
    ]
\\ 
\vspace{1em}

\textcolor{blue}{example\_input} =  \{``image\_id": ``227884", ``width": 444, ``height": 640, ``objects": [``tie.1:[217, 409, 233, 436]", ``tie.2:[212, 409, 233, 507]", ``person.3:[119, 289, 300, 523]"], ``captions": \{``global": ``a man wearing a suit", ``Union(tie.1:[217, 409, 233, 436], tie.2:[212, 409, 233, 507])": ``a purple and black cat sitting on a window ledge", ``Union(tie.2:[212, 409, 233, 507], person.3:[119, 289, 300, 523]) ; Union(tie.1:[217, 409, 233, 436], person.3:[119, 289, 300, 523])": ``a man in a suit and tie sitting at a table with a laptop"\} \} 
};
\end{tikzpicture}
\label{tab:prompt}
\end{table}
 \begin{table}[t]
\centering
\caption{Example of synthesising scene graphs with GPT-4.}
\definecolor{colorbg@intro}{RGB}{240,240,240}     
\resizebox{\textwidth}{!}
{
\begin{tikzpicture}[
    node distance=1cm and 0.5cm,
    auto, 
    description/.style={
        rectangle, rounded corners=2pt, draw, thick, fill=colorbg@intro, 
        text width=\textwidth, 
        align=left, 
        scale=1.0
    },
] 
\node (img1) {\includegraphics[width=0.8in]{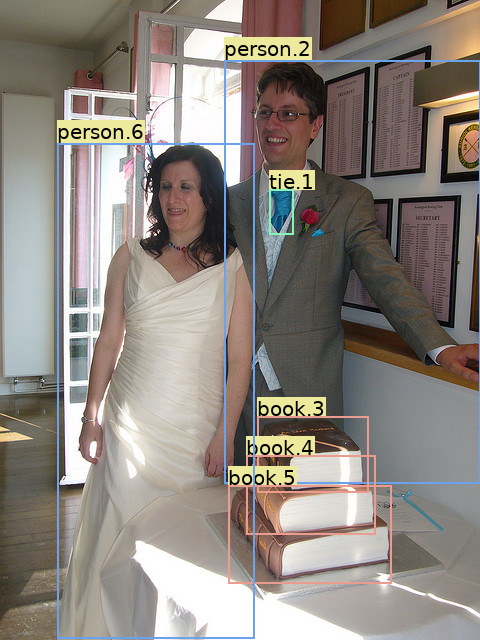}};
\node (input1) [right=of img1, xshift=-5mm, description, scale=0.85] {  
\textcolor{blue}{Example Input} : \{``image\_id": ``395890", ``width": 480, ``height": 640, ``\textcolor{blue}{objects}": [``tie.1:[269, 189, 293, 234]", ``person.2:[224, 60, 480, 483]", ``book.3:[257, 416, 368, 492]", ``book.4:[246, 455, 375, 534]", ``book.5:[228, 485, 391, 583]", ``person.6:[57, 143, 254, 638]"], ``\textcolor{blue}{captions}": \{ ``Union(person.2:[224, 60, 480, 483], book.3:[257, 416, 368, 492])": ``a man and a woman standing next to a cake", ``Union(book.3:[257, 416, 368, 492], book.4:[246, 455, 375, 534])": ``a cake made of books", ``Union(book.3:[257, 416, 368, 492], book.5:[228, 485, 391, 583])": ``a man standing next to a cake that is made of books", ``Union(book.4:[246, 455, 375, 534], book.5:[228, 485, 391, 583])": ``a cake made out of books", ``Union(book.5:[228, 485, 391, 583], person.6:[57, 143, 254, 638])": ``a man and a woman", ``Union(book.4:[246, 455, 375, 534], person.6:[57, 143, 254, 638])": ``a man and a woman standing in front of a cake", ``global ; Union(person.2:[224, 60, 480, 483], person.6:[57, 143, 254, 638]) ; Union(tie.1:[269, 189, 293, 234], person.2:[224, 60, 480, 483]) ; Union(person.2:[224, 60, 480, 483], book.4:[246, 455, 375, 534])": ``a man and woman standing next to a cake"\}\}
};

\node (gpt1) [below=of input1, xshift=-10.65mm, yshift=9mm, description, text width=1.02\textwidth] {
\textcolor{blue}{GPT-4 Response} : 
  \{
    ``image\_id": ``395890",
    ``relationships": [
      \{``source": ``person.2", ``target": ``book.3", ``relation": ``near"\},
      \{``source": ``person.2", ``target": ``person.6", ``relation": ``near"\},
      \{``source": ``person.2", ``target": ``tie.1", ``relation": ``wearing"\}, 
      \{``source": ``person.6", ``target": ``book.4", ``relation": ``near"\},
      \{``source": ``person.6", ``target": ``book.5", ``relation": ``near"\},
      \{``source": ``book.3", ``target": ``book.4", ``relation": ``on"\},
      \{``source": ``book.4", ``target": ``book.5", ``relation": ``on"\}
    ]
  \}
};

\node (user1) [below=of gpt1, description, yshift=9mm,  text width=1.02\textwidth] {
\textcolor{blue}{User}: 
How would you infer that ``person.2" rather than ``person.6" is ``wearing" the ``tie.1" ? 
\vspace{1em} \\ 
\textcolor{blue}{GPT-4}: 
To infer that ``person.2'' rather than ``person.6'' is wearing ``tie.1'', the following observations from the image data can be considered:

\textcolor{red}{Spatial Proximity}: The bounding box of ``tie.1" ([269, 189, 293, 234]) is entirely within the bounding box of ``person.2" ([224, 60, 480, 483]). {This indicates that the tie is located on or very near to person.2}. In contrast, the bounding box of ``person.6" ([57, 143, 254, 638]) does not overlap with that of the tie, suggesting that person.6 is not wearing it. 
\textcolor{red}{Global and Localized Descriptions}: The global description mentions a ``man and woman standing next to a cake,'' but does not specify who is wearing the tie. However, the union of ``tie.1'' and ``person.2'' in the captions suggests a relationship between them. Since ties are typically worn as part of clothing, the logical inference is that person.2 is wearing the tie. 
\textcolor{red}{Logical Consistency}: In typical scenarios, a tie is worn by one person at a time. Given that the tie's location coincides with the location of person.2 and not with person.6, it is logical to conclude that person.2 is the one wearing the tie.
In conclusion, \colorbox{red!30}{the spatial arrangement} of the objects and \colorbox{red!30}{the given descriptions} lead to the inference that ``person.2'' is wearing ``tie.1''.
}; 

\end{tikzpicture}
}
\label{tab:examples}
\end{table}
 
\section{Details of Experiments}
 \subsection{Statistics of Generated Data}
 \label{sec:freq}
\begin{figure}[t]
    \centering
    \begin{minipage}{\textwidth}
         \includegraphics{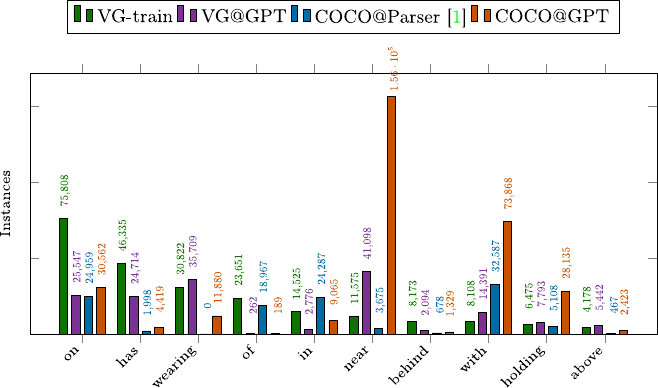}
         \subcaption{Frequency of head-10 predicates.}
    \end{minipage}
        \begin{minipage}{\textwidth}
         \includegraphics{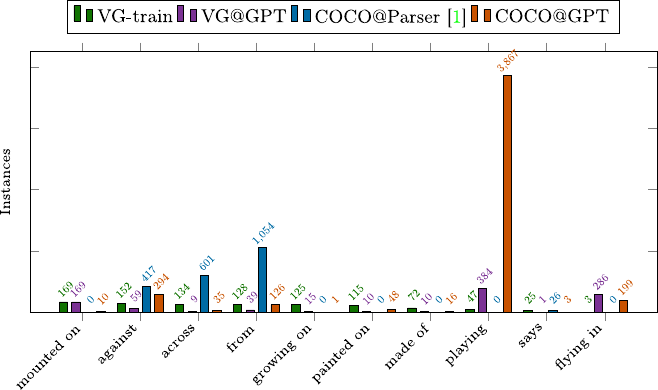}
         \subcaption{Frequencey of tail-10 predicates.}
    \end{minipage}
    \caption{Frequency of head-10 (a) / tail-10 (b) predicates from VG-train, compared across VG@GPT, COCO@Parser\cite{mao2018parser}, and COCO@GPT datasets.}
    \label{fig:top10}
\end{figure}
 
 We provide the statistics of head-10 / tail-10 categories of the VG150 dataset, 
compared across VG@GPT, COCO@Parser, and COCO@GPT data in \cref{fig:top10}.

\subsection{More Experimental Results}
We present the performance of IMP \cite{xu2017scene}, a classic SGG model,  in \cref{tab:exp}. 
From the results, \textit{GPT4SGG} (Row 3) outperforms previous weakly-supervised SGG (Row 2), and fully-supervised SGG (Row 1) in terms of mR@K.

\subsection{Regarding the Lookup Table}
\cref{tab:prompt_lookup} presents the prompt for building a lookup table of semantic matching between two relationship sets. 

\subsection{Training Details of Llama 2}
We use the  trl library \cite{vonwerra2022trl} to fine-tune the Llama 2 - 13B \cite{touvron2023llama} on a 8x NVIDIA A100 (80G) server with the hyper-parameters shown in \cref{tab:hyper}.
\begin{table}[ht]
    \centering
     \caption{Hyper-parameters of fine-tuning Llama 2.}
    \begin{tabular}{cc|cc}
        \toprule
         Value& Item  & Value & Item\\
         \midrule 
         optimizer & AdamW  & batch size& 128 \\
         maximum lr & 1e-4  & warmup steps & 100 \\
         epochs & 5 & lora rank & 128 \\
         lora alpha & 16 & lora dropout & 0.05 \\
         dtype & int8 & & \\ 
         \bottomrule
    \end{tabular}
    \label{tab:hyper}
\end{table}

\subsection{Regarding the Long-tail Bias}
\begin{figure}[t]
    \centering
     \begin{minipage}{\textwidth}
         \includegraphics[width=\textwidth]{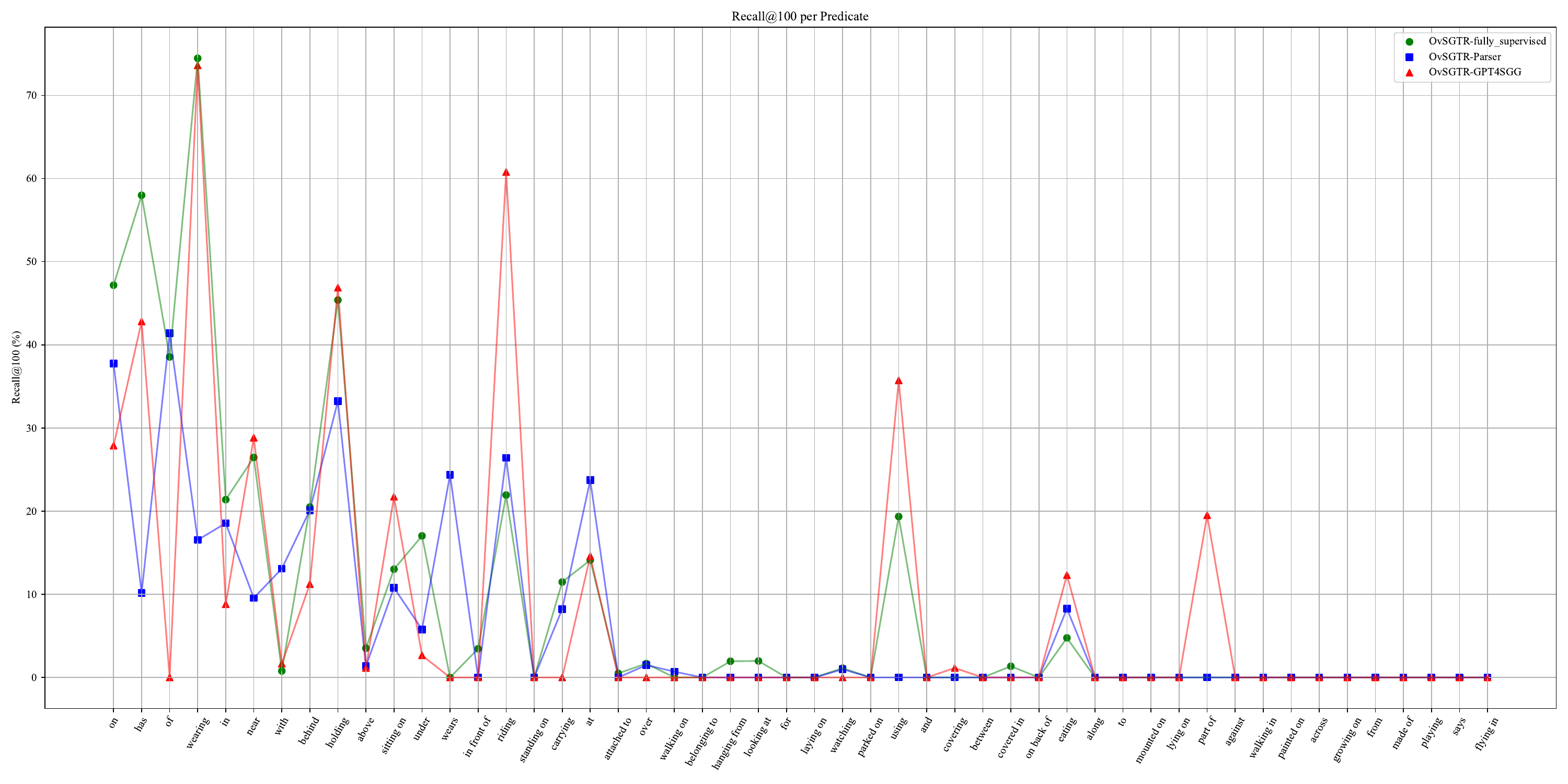}
         \subcaption{Recall@100 per predicate of OvSGTR.}
     \end{minipage}
      \begin{minipage}{\textwidth}
         \includegraphics[width=\textwidth]{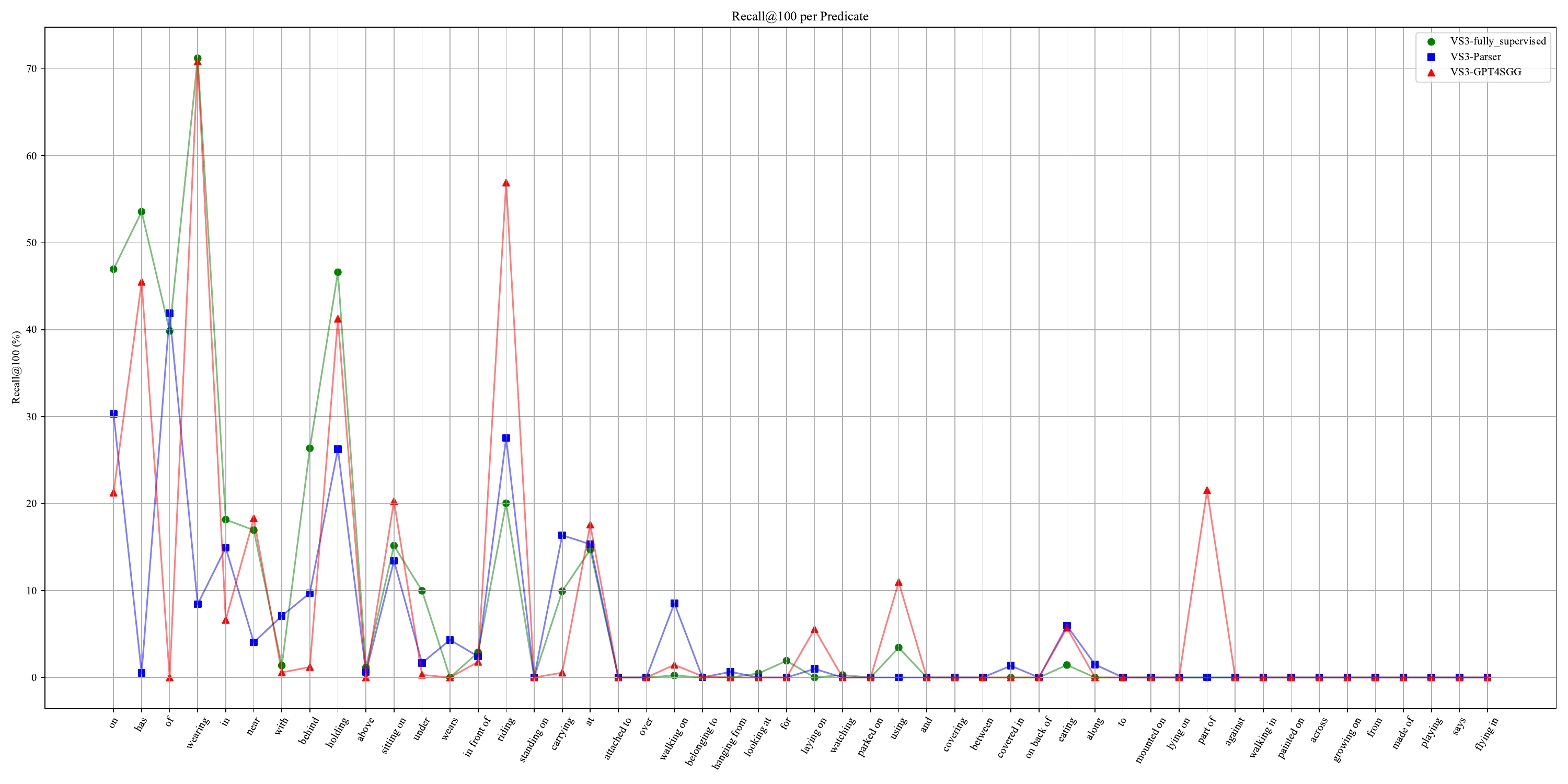}
         \subcaption{Recall@100 per predicate of VS3.}
     \end{minipage}
    \caption{Comparison of three methods (best view via zoom in): fully supervised SGG, parser-based language supervised SGG, GPT4SGG. 
    Models are tested on VG150 test set , and predicates of x-axis are sorted by their frequencies. 
     }
    \label{fig:recall-ovsgtr}
\end{figure}

\cref{fig:recall-ovsgtr} reports Recall@100 per predicate of several models.
The quantitative  results are consistent with the statistics presented in 
 \cref{sec:freq}.
From the results, \textit{GPT4SGG} outperforms parser-based methods except for conjunction words like ``on'' and ``of''.
For actions, parser-based methods obtain a better recall of ``wears'' due to the lemmatization in parsing triplets (\eg, ``wearing'' would be lemmatized as
``wears''). 
Moreover, \textit{GPT4SGG} surpasses fully-supervised methods in predicates like 
``riding'', ``near'' and  ``part of''. 
 
\subsection{Compare with GPT-4V}
Given the expensive cost of GPT-4V API (\url{https://openai.com/pricing}), we randomly choose 100 images from the VG150 validation set to test the capability of GPT-4V, as shown in \cref{tab:recall_4v}. 
Images are resized to no longer than 512, and object information, including names and bounding boxes, is provided.
\begin{table}[t]
    \centering
    \caption{Comparison with GPT-4V on 100 images of VG150 validation set.} 
    \resizebox{0.9\textwidth}{!}{
    \begin{tabular}{c|c|c|c|c}
        \toprule 
        \multirow{2}{*}{Model} & \#Triplets   & Recall  & Token Usage  &  {Cost} \\
          & (GT/Generated)  &(original/\textit{w.} lookup) & (Context/Generated)&   (USD)\\
        \midrule 
         GPT4SGG & 558/610 & 17.5/18.1 & 91k/17k& ${\sim}$1.4 \\
         GPT-4V &  558/1145  & 17.6/19.6& 56k/30k & ${\sim}$1.5  \\ 
        \bottomrule
    \end{tabular}
    }
    \label{tab:recall_4v}
\end{table}
Without surprising, GPT-4V achieves a higher recall with the help of original image input. 
However, GPT-4V is a black box, and \textit{GPT4SGG} can be customized with local LLMs, achieving comparable performance.

\begin{table}[t]
\centering
\definecolor{colorbg@intro}{RGB}{240,240,240}     
\caption{Prompt for building a lookup table with GPT-4. 
Set A is the set to be matched with Set B (VG150 relationships). 
}
\begin{tikzpicture}[
    node distance=1cm and 0.5cm,
    auto, 
    description/.style={
        rectangle, rounded corners=2pt, draw, thick, fill=colorbg@intro, 
        text width=\textwidth, 
        align=left, 
        scale=1.0
    },
] 
\node[description] {  
\textcolor{blue}{messages} = [
    \{ \verb|"|\textbf{role}\verb|"|: \verb|"|user\verb|"|, \verb|"|\textbf{content}\verb|"|:
    
\verb|"""|Given the relationships in Set A:
\verb|`|
\{\textcolor{blue}{Set A}\}
\verb|'|,   and Set B:
\verb|`|
["above", "across", "against", "along", "and", "at", "attached to", "behind", "belonging to", "between", "carrying", "covered in", "covering", "eating", "flying in", "for", "from", "growing on", "hanging from", "has", "holding", "in", "in front of", "laying on", "looking at", "lying on", "made of", "mounted on", "near", "of", "on", "on back of", "over", "painted on", "parked on", "part of", "playing", "riding", "says", "sitting on", "standing on", "to", "under", "using", "walking in", "walking on", "watching", "wearing", "wears", "with"]
\verb|'|,
Please associate the two sets based on semantic comparison. \\
1) We should visit each item of set A only once and find the associated item in set B. \\ 
2) Use a direction flag where 1 indicates the same meaning, 2 indicates a weakly similar meaning,  -1 indicates an antonym, and 0 refers to no associated item. \\ 
3) Format the result as a dictionary with keys "source", "target", "direction" . "source" must be from set A ,  and "target" must be from set B or None if there is no association. \\ 
4) Provide the complete associations as a list of dicts in JSON format.

Example output: \\ 
``` \\ 
\verb|[| 
\{"source": "on", "target": "on", "direction": 1\}, \\ 
\{"source": "next to", "target": "near", "direction": 1\}, \\ 
\{"source": "ridden by", "target": "riding", "direction": -1\} 
\verb|]| \\ 
``` \\ 

\verb|###| Output for Set A: \verb|"""|
\}
] 
};
\end{tikzpicture}
\label{tab:prompt_lookup}
\end{table}

\begin{table}[t]
\centering
\definecolor{colorbg@intro}{RGB}{240,240,240}     
\caption{Prompt for synthesizing scene graphs using GPT-4V. 
}
\begin{tikzpicture}[
    node distance=1cm and 0.5cm,
    auto, 
    description/.style={
        rectangle, rounded corners=2pt, draw, thick, fill=colorbg@intro, 
        text width=\textwidth, 
        align=left, 
        scale=1.0
    },
] 
\node[description] {  
\textcolor{blue}{messages} = [\{
    \{ \verb|"|\textbf{role}\verb|"|: \verb|"|user\verb|"|, \verb|"|\textbf{content}\verb|"|:
    [\{
    \verb|"|type\verb|"|: \verb|"|text\verb|"|, 
    \verb|"|text\verb|"|: f\verb|"""| 
    Given a set of detected objects in an image, each object is characterized by a name, a bounding box in \verb|"|xyxy\verb|"| format, and a confidence score (optional). Please generate a scene graph to describe this image.  Example Output: \\ 
\{
 \verb|"|relationships": [
  \{\verb|"|source\verb|"|: \verb|"|object\_id1\verb|"|, \verb|"|target\verb|"|: \verb|"|object\_id2\verb|"|, \verb|"|relation\verb|"|: \verb|"|relation\_type\verb|"|\},
  ... 
 ]
\}\\ 
Now, objects are \{\textcolor{blue}{objects}\}; 
the original width and height of the provided image are \{\textcolor{blue}{size}\}. Please output the scene graph in JSON style without any comments. \verb|"""|
\}, \\
\{ \verb|"|type\verb|"|: \verb|"|image\_url\verb|"|,
\verb|"|image\_url\verb|"|: \{
    \verb|"|url\verb|"|: f\verb|"|data:image/jpeg;base64,\{\textcolor{blue}{base64\_image}\}\verb|"|
\}
] 
\vspace{1em}

An example of \textcolor{blue}{objects} =  [``tie.1:[217, 409, 233, 436]", ``tie.2:[212, 409, 233, 507]", ``person.3:[119, 289, 300, 523]"]  
};
\end{tikzpicture}
\label{tab:prompt_4v}
\end{table}

\clearpage  

%
%
\bibliographystyle{splncs04}
\bibliography{egbib}
\end{document}